\documentclass[twoside]{article}
\usepackage{PRIMEarxiv}
\usepackage[utf8]{inputenc} 
\usepackage[T1]{fontenc}    
\usepackage{hyperref}       \usepackage{url}            
\usepackage{booktabs}       
\usepackage{amsfonts}       
\usepackage{nicefrac}       
\usepackage{microtype}      
\usepackage{listings}
\usepackage{fancyhdr}       
\usepackage{graphicx}       
\graphicspath{{media/}}     
\usepackage{subcaption}
\usepackage{calc}
\usepackage{amssymb}
\usepackage{amstext}
\usepackage{amsmath}
\usepackage{amsthm}
\usepackage{multicol}
\usepackage{pslatex}
\usepackage{apalike}
\usepackage{algorithm2e}
\usepackage[bottom]{footmisc}
\usepackage{multirow}
\usepackage{xcolor}
\usepackage{verbatim}

\title{Corpora deduplication or duplication in Natural Language Processing of few resourced languages ? A case of study: The Mexico's Nahuatl
\thanks{\textit{\underline{Citation}}: 
\textbf{Guzman-Landa et al. Corpora deduplication or duplication in Natural Language Processing of few resourced languages ? A case of study: The Mexico's Nahuatl, ArXiv cs.CL 8 pages.}} 
}

\author{Juan-José Guzm\'an-Landa$^1$, Juan-Manuel Torres-Moreno$^{1,2}$, Graham Ranger$^3$\\
\{$^1$LIA; $^3$LCTT\}, Université d'Avignon. Avignon, France\\
$^2$LIFAT, Université de Tours. Tours, France\\
\texttt{\{juan-jose.guzman-landa, juan-manuel.torres, graham.ranger\}@univ-avignon.fr}
\AND
Miguel Figueroa-Saavedra$^5$, 
Martha-Lorena Avenda\~no-Garrido$^4$\\ 
$^4$Facultad de Matem\'aticas, $^5$Instituto de Investigaciones en Educación,\\
Universidad Veracruzana, Xalapa, Mexico; 
\texttt{\{migfigueroa, maravendano\}@uv.mx} 
\AND
Elvys Linhares-Pontes$^7$ \\
$^7$Trading Central, France; \texttt{elvys.linharespontes@tradingcentral.com}
\AND Luis-Gil Moreno-Jiménez$^6$\\ 
$^6$Independent Researcher, France; 
\texttt{lmoreno30470@gmail.com}
}

\begin{document}

\maketitle

\fancyhead[CE]{\textsc{\footnotesize{J.J. Guzman-Landa et al.}}}
\fancyhead[CO]{\textsc{\footnotesize{Corpora deduplication or duplication in NLP of few resourced languages ? The Nawatl of Mexico}}}

\begin{abstract}
In this article, we seek to answer the following question: could data duplication be useful in Natural Language Processing (NLP) for languages with limited computational resources? In this type of languages (or $\pi$-languages) \cite{these-pi,abdillahi:hal-01311495}, corpora available for training Large Language Models are virtually non-existent. In particular, we will study the impact of corpora expansion in Nawatl, an agglutinative and polysynthetic $\pi$-language spoken by over 2 million people, with a large number of dialectal varieties \cite{Lastra1986areas,Canger1988NahuatlDA}. The aim is to expand the new $\pi$-{\sc yalli} corpus, which contains a limited number of Nawatl texts \cite{piyalliTALN}, by duplicating it in a controlled way. In our experiments, we will use the incremental duplication technique. The aim is to learn embeddings that are well-suited to NLP tasks. 
Thus, static embeddings were trained and evaluated in a sentence-level semantic similarity task \cite{2025cfgnahuatlautomaticcorpora}. Our results show a moderate improvement in performance when using incremental duplication compared to the results obtained using only the corpus without expansion. Furthermore, to our knowledge, this technique has not yet been used in the literature.
\end{abstract}

\vspace*{14pt}{\textbf{Key words:}} \begin{footnotesize}{Nawatl; Corpora expansion; Large Language Models; Sentence Semantic Similarity.}\end{footnotesize}

\section{Introduction}

Large Language Models (LLM) require corpora containing vast amounts of textual data in order to capture the deep structure of a language. 
These amounts often run into the hundreds of millions or even billions of words. 
Furthermore, it has been found that performance increases logarithmically with corpus size \cite{kaplan2020scalinglawsneurallanguage}. 
This massive data requirement 
poses a major problem for the development of LLMs for languages with few computational resources, or $\pi$-languages, as opposed to $\tau$-languages or languages with abundant resources \cite{these-pi,abdillahi:hal-01311495}. Indeed, $\pi$-languages suffer from a severe lack of textual corpora that are both representative and large-scale, making it impossible to train LLMs. Consequently, these languages remain under-represented in Natural Language Processing (NLP), perpetuating a linguistic bias that limits their usefulness for the communities that speak them.

One example of the Americas' $\pi$-languages is Nawatl (or Nahuatl), one of Mexico’s indigenous national languages.
In this country, Nawatl has been recognised as the second national language, after Spanish, with approximately 1.65 million Nawatl speakers \cite{inegi2020censo}. 
This language exhibits great dialectal diversity, with 29 recognised varieties spread across four major regions in Mexico: Western, Central, Eastern and Huasteca\footnote{See Ethnologue, 2025\footnote{\url{https://www.ethnologue.com}} and \cite{Lastra1986areas}}. 
This linguistic diversity poses enormous challenges for the development of NLP tools, as it involves correctly handling significant variations in spelling and lexical choices \cite{zimmerman,olko2016bridging,hansen2024nahuatl}.
To this end, a symbolic unifier for Nawatl spellings has recently been proposed \cite{MICAI-piyalli-unigraph}. Although the publication of digital content in Nawatl is constantly increasing, its  dispersion and great dialectal diversity have not allowed for a clear presence and accessibility to be established within the few available corpora. 
Nevertheless, the availability of digital Nawatl documents and their written use are essential for the current revitalisation of the language \cite{pugh-etal-2025-ihquin}.

Our approach to addressing the problem of a lack of corpora involves the controlled duplication of available textual data. Combined with other techniques, this strategy could serve as a basis—in the case of $\pi$-languages—for expanding corpora on a larger scale. These corpora, in turn, could be used for training contextualised or static LLMs \cite{tunstall2022transformers,goyal2018deep}.
More specifically, our objective is to sufficiently expand the Nawatl $\pi$-\textsc{yalli} corpus\footnote{This corpus is available on the website: \url{https://demo-lia.univ-avignon.fr/pi-yalli}}, which should have a positive impact on learning models producing static embeddings.

The structure of the paper is as follows: 
Section \ref{sec:etat_de_lart} provides a review of corpus expansion techniques in under-resourced languages.
Section \ref{sec:nawatl} introduces the Nawatl language and the $\pi$-{\sc yalli} corpus.
Section \ref{sec:duplicacion} presents the incremental duplication technique. 
Section \ref{sec:experiments} presents experiments with the expanded corpus in the context of a semantic similarity task. Finally, section \ref{sec:conclusions_future} concludes the paper and suggests avenues for future research.

\section{State-of-the-art}
\label{sec:etat_de_lart}

In the literature, duplicate data is primarily addressed as a problem in languages rich in  computational resources (or $\tau$-languages) \cite{abdillahi:hal-01311495}, rather than as a technique for increasing corpus size.  Indeed, the vast amount of data available on the internet leads to significant redundancy in collected corpora, which hinders model training \cite{lee-etal-2022-deduplicating,penedo2024fineweb}. Consequently, most research focuses on the detection and removal of duplicates (or deduplication), particularly in the context of corpora intended for LLMs training.

This problem is particularly pronounced in $\tau$-languages, for which the volume of available textual data is quite substantial, but also highly redundant. For this reason, most research aims to produce large-scale corpora whilst minimising duplicates as much as possible. Thus, FineWeb \cite{penedo2024fineweb} and CCNet \cite{wenzek-etal-2020-ccnet} demonstrate filtering and deduplication techniques to produce high-quality, non-redundant corpora.

However, the situation is different in the case of $\pi$-languages, where the problem is not an excess of data, but a scarcity of it. In this context, data augmentation (DA) could be a promising strategy for expanding currently available corpora to compensate for the lack of resources \cite{feng-etal-2021-survey,chen-etal-2023-empirical}. The data augmentation techniques proposed in the literature can be classified into two main approaches: 

\begin{description}
    \item [Lexical level.] The EDA (Easy Data Augmentation) method \cite{wei-zou-2019-eda} performs simple operations such as synonym substitution, as well as the insertion, deletion or random replacement of words. It has been applied solely in the context of text classification, and the results show performance ranging from 87.8\% to 88.6\%, representing improvements of less than 1\%. These techniques use dictionaries to work with synonyms. EDA has not been applied to $\pi$-languages.
    
    \item [Syntagmatic level.] For languages lacking dictionaries, there are techniques such as EDDA (Easy Distributional Data Augmentation) and TSSR ({Type Specific Similar word Replacement}) \cite{mahamud2023distributionaldataaugmentationmethods}, which utilise distributional context and morphosyntactic labels to address this shortcoming. TSSR requires the data to be annotated with POS\footnote{\textit{Part-Of-Speech}.} tags. EDDA relies on the latent space generated by Word2Vec instead of a dictionary. These techniques have been used on Swedish corpora.
\end{description}

In this article, we propose an approach aimed at incrementally duplicating a corpus identically, without the need for lexical resources such as POS taggers or dictionaries. In the case of Nawatl, these resources (which are sometimes non-existent) are difficult to use directly due to the language’s high degree of agglutination and poly-synthesism. Furthermore, the few dictionaries available do not cover all dialectal varieties. Finally, we believe that EDA-type techniques and their random mechanism may introduce syntactic and semantic biases. For these reasons, our aim is to avoid them in our proposal.

\section{The Nawatl language and the $\pi$-{\sc yalli} corpus}
\label{sec:nawatl}

\subsection{Nawatl language}

The Nawatl language is a polysynthetic and agglutinative language.
In other terms, to form words and convey meaning, it attaches various morphemes to a nominal or verbal root. At the syntactic level, Nawatl sentences follow a basic verb–subject–object (VSO) word order, although this can be flexible. Thus, there are VO, VS, VOS and, less frequently, SV, SVO and SOV word orders \cite{2025cfgnahuatlautomaticcorpora}, which meet the needs of speakers. Furthermore, the syntactic and semantic relationships between words and clauses are established through the valency of the verb and the use of conjunctive particles. These particles may also function as markers and discourse connectors. 

Another distinctive feature of Nawatl is that words can be regarded as complete sentences, and this is particularly true of verbal words.  We therefore refer to them as phrase-words or ‘single-word phrases’, as their morphology includes the subject and predicate, as well as information on the actants, and modal, directional and relational elements \cite{Launey1978introduction,Wright-Carr2016,floresnajera2019gramatica,sasakidivide}.

Given its oral nature, there are very few written resources available for this language\footnote{See \cite{piyalliTALN} on this subject.}. Combined with the lack of standardised writing systems, this makes automated processing extremely difficult.

\subsection{Available Resources and the $\pi$-{\sc yalli} corpus}

There are very few tools and resources available for Nawatl language.
To our knowledge, only one machine translation tool has been available for the Huasteca variety\footnote{See \textit{Google Translate} (\url{https://translate.google.com.mx/?hl=es&sl=nhe&tl=es&op=translate})} since 2024. 
Meanwhile, in 2017, the  {\it Instituto de Ingenier\'ia} at the {\it Universidad Nacional Aut\'onoma  de México} (UNAM) launched \textit{Axolotl}, a corpus of bilingual Spanish/Nawatl documents\footnote{The \textit{Axolotl} corpus is available at the following address: \url{http://www.corpus.unam.mx/axolotl}}. 
Furthermore, a spelling unifier \cite{MICAI-piyalli-unigraph} and the new $\pi$-\textsc{yalli} corpus have recently been introduced in France.
However, many varieties and texts remain inaccessible.
This has a negative impact on the development of machine learning-based tools, thereby preventing widespread use and adoption by Nahua-speaking communities. 

The Nawatl $\pi$-{\sc yalli} corpus \cite{piyalliTALN} is a resource available for machine learning and NLP algorithms. It is a heterogeneous corpus in terms of topics (around 20) and dialectal varieties (around 25) of Nawatl, spoken mainly in Mexico and El Salvador.
It contains a limited number of words (around 6.6 million) and sentences, but it has been used successfully in various NLP tasks \cite{piyalliTALN,2026classifyingdialectalnawatlvarieties}.
It is, however, useful for training language models such as statistical or vector models: TF-IDF \cite{Manning:99}, BM25 \cite{robertson2004simple}, TF-PDF \cite{bun2002topic} or static embedding models such as Word2Vec \cite{Mikolov2013distributed}, FastText \cite{bojanowski-etal-2017-enriching} or Glove \cite{glove}, 
but clearly unsuitable for training contextualised vector models using BERT-style transformers \cite{bert_04805}. 

Indeed, it has been reported that contextual LLMs require between 10 and 100 million tokens to obtain stable embeddings~\cite{micheli2020importancepretrainingdatavolume}.
This is why we decided to expand the $\pi$-\textsc{yalli} corpus by duplicating it in a controlled manner in order to assess the impact of this technique on learning algorithms.

\color{black}

\section{Corpora Duplication}
\label{sec:duplicacion}

Given the scarcity of computational resources available for $\pi$-languages, we adopted a specific strategy: to incrementally augment the size of the available corpus by reusing the same textual content.
At first glance, such a strategy might appear to have no positive impact on the training of embeddings. 

In fact, it would run against current recommendations.
Indeed, it has been found that corpus deduplication is a crucial step in achieving effective embedding learning \cite{lee-etal-2022-deduplicating}.
This is particularly true in the case of languages with extensive computational resources, or $\tau$-languages, where there is no need to duplicate corpora. Indeed, sentences that appear 60,000 times or more pose a real problem for dense word representation, as they often lead to over-fitting in neural models. 

In addition to the lack of resources, it must be considered that Nawatl is an agglutinative language and, as a result, the frequent use of phrase-words reduces the number of what we normally understand as ‘words’, compared to other types of languages.
That is to say, what in non-agglutinative languages might be expressed in five or six words is expressed in a single word in Nawatl.
This phenomenon is very evident in translation. All of this has consequences for the number of words (tokens) available in the corpora.
However, our hypothesis is that a \textit{controlled} increase in the number of occurrences could facilitate the learning of embeddings in the case of $\pi$-languages, and in particular we have studied this in the case of Mexican Nawatl.

We decided to test our hypothesis regarding the impact of corpora expansion on learning algorithms through empirical means.
The aim, of course, is to have a positive impact on the learning of static word embeddings.

To do this, we duplicated the corpus $\pi$-{\sc yalli} $\rho$ times, creating identical copies; where $\rho$ = [1, 2, 4,..., 26, 28, 30] times its original size. We thus incrementally generated corpora of approximately: 6.6 million (original), 13.2 million, 19.8 million, ..., 198 million words (duplicated 30$\times$ times).
These quantities represent a considerable volume of text, enabling the size of the $\pi$-{\sc yalli} corpus to be significantly expanded.
The corpus data has been processed through data cleaning, paragraph and sentence segmentation, and the removal of some stopwords \cite{2025cfgnahuatlautomaticcorpora}.

The experiments on a semantic task and the results relating to this duplication strategy will be detailed in the following section.

\section{Experiments} 
\label{sec:experiments}

Semantic similarity, a classic task in NLP, involves evaluating various models (statistical, neural networks, etc.) using appropriate evaluation protocols \cite{francis-landau-etal-2016-capturing}. 
In our study, the aim is to calculate the semantic similarity between reference sentences and sets of candidate sentences, which may be semantically close to or distant from the references. This results in rankings of the candidate sentences, which will be compared to rankings produced by native Nawatl speakers, via a statistical estimator.

\subsection{Semantic Similarity Task using static embeddings}

In this study, we adopted the same semantic evaluation protocol used in the literature \cite{piyalliTALN}:
30 reference sentences and a set of 5 candidate sentences per reference to be semantically ranked.
This allows us to estimate the impact of incremental duplication on embedding learning and also to measure their quality on a semantic proximity task.

Static embeddings have been widely used in NLP tasks (classification, analogies, semantic similarity, etc.), but they have been replaced by transformer-type contextual embedding models (such as the BERT model), whose popularity is due to their excellent performance \cite{bert_04805}. Although transformers have demonstrated their superiority in NLP tasks, this has only been possible in $\tau$-languages, which are rich in computational resources.
Indeed, these types of models require large amounts of text data to train effectively.
The situation changes completely when it comes to processing $\pi$-languages. 
Under this scenario, static embeddings are still competitive, as they can be generated from scratch, are quick to train and, most importantly, non-contextualised models require very small corpora to learn effectively.

Among the most popular algorithms for generating static embeddings are: Word2vec \cite{mikolov2013efficientestimationwordrepresentations}, FastText \cite{bojanowski-etal-2017-enriching} and Glove \cite{pennington-etal-2014-glove}.
We therefore trained and compared them for word embedding learning on duplicate, large-scale corpora. 
The quality of the embeddings obtained was assessed by using them in a semantic similarity task involving Nawatl sentences. This established a ranking. The similarity between the rankings was estimated using Kendall’s $\tau$ coefficient. Kendall’s $\tau$  is a suitable non-parametric measure of correlation that assesses the ordinal association between two variables, i.e. the degree of agreement between two rankings \cite{665905b2-6123-3642-832e-05dbc1f48979}.

\color{black}

\subsection{Results and Discussion}
\label{sec:result_duplicacion}

The results obtained enable us to assess the performance of the static models trained on identical duplicate corpora. 

Firstly, we observed a significant difference between the CBow and SkipGram versions of the FastText and Word2Vec algorithms.
According to the literature, CBow is an architecture that focuses on predicting an unknown word X based on its context (the set $E(X)$ of words surrounding X). 
X is then predicted based on the information provided by the context $E$. On the other hand, in Skipgram mode, it is the context (the words surrounding X) that is predicted based on the word X. Furthermore, the Word2Vec architecture generates one vector (embedding) per word in the vocabulary, whereas FastText generates one embedding per $n$-gram of characters (token). This allows vectors containing more information to be constructed thanks to the $n$-grams. For these reasons, FastText in Skipgram mode generally achieves better results than Word2Vec. We have corroborated this in our experiments, where the FastText Skipgram versions perform significantly better than the others.

\begin{figure}[h!]
\centering
\includegraphics[width=0.9\linewidth]{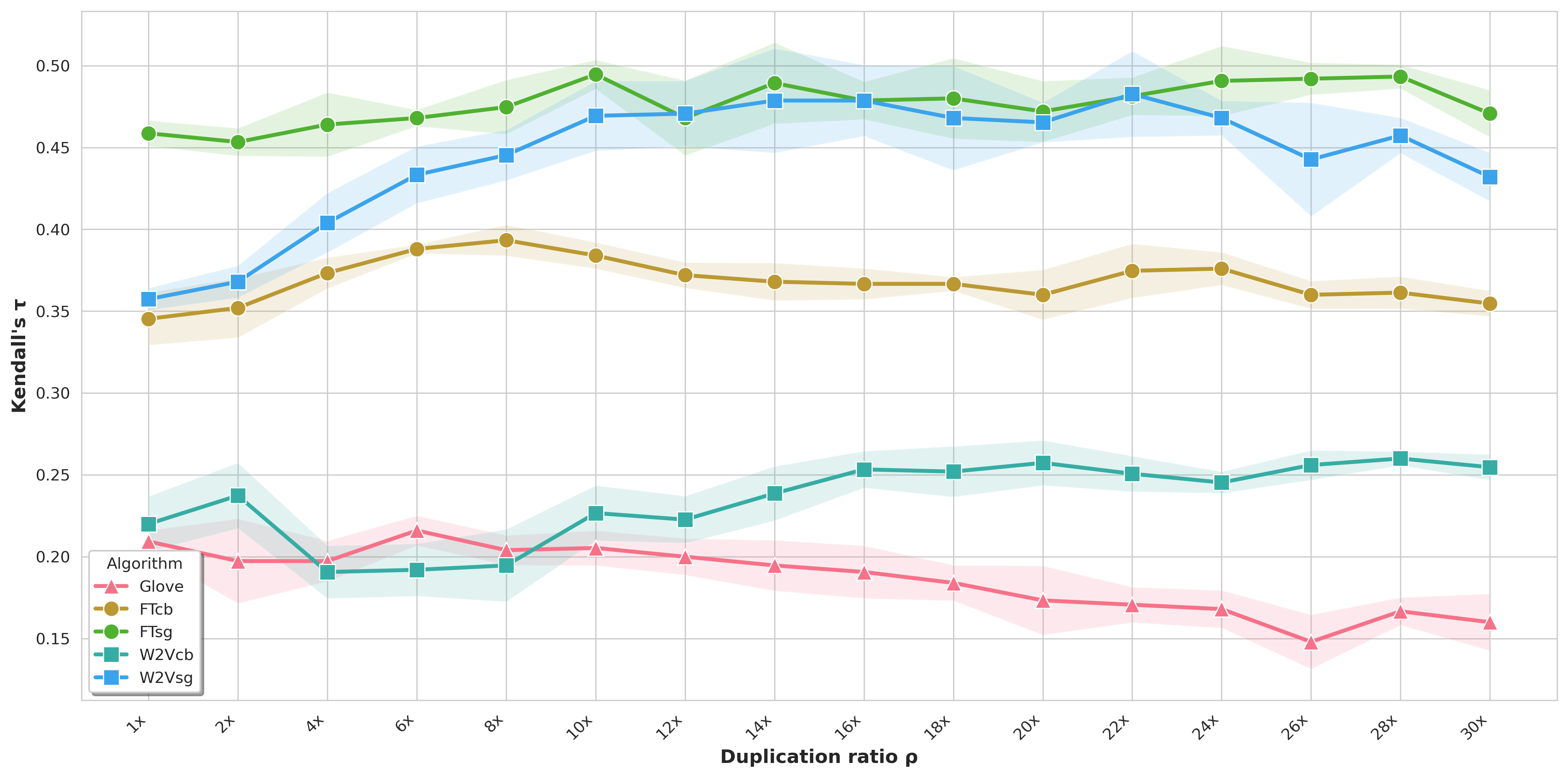}
  \caption{Sentence semantic similarity task: Kendall's coefficient $\langle\tau\rangle$.
  Learning on incrementally duplicated corpora $\pi$-{\sc yalli}, using three static models: Glove, FastText (Cbow=FTcb, Skipgram=FTsg) and Word2vec (Cbow=W2Vcb, Skipgram=W2Vsg). We have performed five runs per duplication point $\rho$.}
  \label{fig:naw_duplication_tous_modeles}
\end{figure}

Figure \ref{fig:naw_duplication_tous_modeles} illustrates our results.
Each curve shows the performance of models trained on corpora ranging in size from 1$\times$ (without duplication) to 30$\times$ duplications.
Each point represents the mean Kendall's $\tau$ across 5 runs, and each band represents their standard deviation. The FastText algorithm in Skipgram mode achieves the best performance in most runs.

However, it must be said that Word2Vec, also in Skipgram mode, benefited significantly from the proposed duplication technique, with steady improvements in rank as the number of duplications increased from 1$\times$ to 16$\times$.
In contrast, the Glove algorithm generally stagnated and produced rather disappointing results as the number of duplications increased.

Further details of our results are shown in Table (\ref{tab:tau_kendalls}): the maximum values of $\tau$, the improvement and the learning time. With the exception of Glove —which achieves only a small improvement— exact replication yields moderate or significant benefits. 
We have observed that Word2Vec performs better when the duplication ratio $\rho=[20,22]$.
On the other hand, FastText achieves its maximum values when $\rho=[8,10]$.
This shows that FastText, with a lower duplication ratio $\rho$, generates higher-quality static embeddings. 

To compare our results using the same algorithms but trained on different corpora – without duplication – we used three commonly available pre-trained embedding models: 
\begin{enumerate}
\item FastText trained on the Common Crawl corpus\footnote{https://commoncrawl.org/} ;
\item FastText  trained on  Nawatl Wikipedia corpus\footnote{FastText has been trained on 157 languages; see the website: \url{https://fasttext.cc/docs/en/crawl-vectors.html}}  ;
\item Word2Vec also trained on  Nawatl Wikipedia corpus\footnote{Available on the website: \url{https://sparknFastText has been trained on 157 languages; see the website:lp.org/2022/03/16/w2v_cc_300d_nah_3_0.html}}.
\end{enumerate}
Their results are shown at the bottom of the Table (\ref{tab:tau_kendalls}).

\begin{table}[ht]
\centering
\begin{tabular}{|c|c|c|r|c|c|}
\hline
\bf Model & \bf $\tau$ (1$\times$) & \bf max $\tau$ ($\rho\times$) & $\rho$ & \bf Improvement \% &\bf Time (min) \\ \hline 
FastText Skipgram     & 0.459 & 0.495 & 10 & 7,8 & 46,6 \\
Word2Vec Skipgram     & 0.357 & 0.483 & 22 & 35,3 & 39,3 \\
FastText Cbow         & 0.345 & 0.393 & 8  & 13,9 & 43,7 \\
Word2Vec Cbow         & 0.220 & 0.257 & 20 & 16,8 & 14,9 \\
Glove                 & 0.209 & 0.216 & 6  & 3.4 & 6.5 \\ \hline
FastText/Wikipedia    & 0.242 & - & - & - & -\\
FastText/Common Crawl & 0.240 & - & - & - & -\\
Word2Vec/Wikipedia    & 0.240 & - & - & - & -\\
\hline
\end{tabular}
\caption{The average Kendall's $\tau$ across five runs of the models. 1$\times$: without duplication, and the maximum $\tau$ obtained with $\rho\times$ duplications of the $\pi$-{\sc yalli} corpus. \% = percentage improvement in $\tau$. The time is approximate for a single run relative to $\rho$. The runs were carried out on a cluster of machines, with a requirement of [8, 12] cores and [16, 64] GB of RAM, running under GNU/Linux in SLURM  (\textit{Simple Linux Utility for Resource Management}) mode.}
\label{tab:tau_kendalls}
\end{table}

\section{Conclusions and Future Works}
\label{sec:conclusions_future}

Although the results are not spectacular, the technique of corpora duplication applied on agglutinative languages with few resources seems to facilitate the training of certain models based on static embeddings. 
Indeed, the embeddings trained in this way capture the structure of the language even more effectively. 
In particular, exact duplication $\rho\times$ times has enabled us to significantly outperform the results obtained with the authentic $\pi$-\textsc{yalli} corpus without duplication.

Indeed, we have shown that controlled corpora duplication from its original size enabled the FastText and Word2Vec algorithms, in their Skipgram mode, to improve their performance in this semantic similarity task.
This was measured using Kendall’s average $\tau$, which increased from $\tau$=0.459 to $\tau$=0.495 (an increase of approximately 8\%) when using the FastText algorithm.
On the other hand, Word2Vec Skipgram improved from $\tau$=0.357 to $\tau$=0.483, i.e. an increase of over 35\%.
In contrast, the Glove algorithm showed, overall, a drop in performance that remains to be explained.

In future work, we will explore the normalisation of textual data and its impact on duplication techniques.
We therefore plan to investigate different ways of expanding the Nawatl $\pi$-{\sc yalli} corpus: either through various types of normalisation, or through duplications focused on specific subjects and dialectal varieties.
Similarly, we intend to evaluate the contribution of the expanded corpora to the tasks of Automatic Text Summarisation \cite{torres14} and Named Entity recognition —such as place names— in Nawatl.

\section*{Acknowledgments}

This research was funded by an Intermedius PhD grant from Université d'Avignon (AU France), and partially funded by grants from the Laboratoire Informatique d'Avignon (LIA) and the Agorantic Research Federation (AU France).

\bibliographystyle{apalike}
\bibliography{biblio}

\end{document}